\documentclass[conference]{IEEEtran}
\usepackage{times}

\usepackage{float} 
\usepackage{bbm}
\usepackage{dirtytalk}
\usepackage[noend]{algpseudocode}
\usepackage{algorithmicx} 
\usepackage{algorithm}
\usepackage{amsmath,amssymb,amsthm,mathtools} % math
\usepackage{graphicx}
\usepackage{subfigure}
\usepackage{caption}
\usepackage{subcaption}   
\usepackage{soul}
\let\labelindent\relax
\usepackage{stfloats,microtype,balance,enumitem}
\usepackage[some,bottom]{background}

\usepackage[numbers]{natbib}
\usepackage{multicol}
\usepackage[bookmarks=true]{hyperref}

\newcommand{\SUBMIT}
\begin{document}

\newcommand{\waypoint}[1]{\ensuremath{\textnormal{WP}_{#1}}}
\newcommand{\waypoints}{\ensuremath{\textnormal{WPs}}}

% paper title
% \title{Possible Title: Eliminating Collisions with New Obstacles: A Preliminary study}

\title{Towards Online Safety Corrections for Robotic Manipulation Policies}

% You will get a Paper-ID when submitting a pdf file to the conference system
% \author{Author Names Omitted for Anonymous Review. Paper-ID [14]}

\author{\authorblockN{Ariana Spalter}
\authorblockA{U.S. Naval Research Laboratory\\
  Washington D.C.
  United States\\
Email: ariana.y.spalter.civ@us.navy.mil}
\and
\authorblockN{Mark `Mak' Roberts}
\authorblockA{U.S. Naval Research Laboratory \\
  Washington D.C. 
  United States \\
Email: mark.c.roberts20.civ@us.navy.mil}
\and
\authorblockN{Laura M. Hiatt }
\authorblockA{U.S. Naval Research Laboratory \\
  Washington D.C.
  United States\\
Email: laura.m.hiatt.civ@us.navy.mil}}

% avoiding spaces at the end of the author lines is not a problem with
% conference papers because we don't use \thanks or \IEEEmembership

% for over three affiliations, or if they all won't fit within the width
% of the page, use this alternative format:
% 
%\author{\authorblockN{Michael Shell\authorrefmark{1},
%Homer Simpson\authorrefmark{2},
%James Kirk\authorrefmark{3}, 
%Montgomery Scott\authorrefmark{3} and
%Eldon Tyrell\authorrefmark{4}}
%\authorblockA{\authorrefmark{1}School of Electrical and Computer Engineering\\
%Georgia Institute of Technology,
%Atlanta, Georgia 30332--0250\\ Email: mshell@ece.gatech.edu}
%\authorblockA{\authorrefmark{2}Twentieth Century Fox, Springfield, USA\\
%Email: homer@thesimpsons.com}
%\authorblockA{\authorrefmark{3}Starfleet Academy, San Francisco, California 96678-2391\\
%Telephone: (800) 555--1212, Fax: (888) 555--1212}
%\authorblockA{\authorrefmark{4}Tyrell Inc., 123 Replicant Street, Los Angeles, California 90210--4321}}

\maketitle

\SetBgContents{\begin{small}\textbf{DISTRIBUTION STATEMENT A.} Approved for public release. Distribution unlimited.\end{small}}
\SetBgScale{1}
\SetBgOpacity{1}
\SetBgVshift{1cm}
\SetBgColor{black}
\BgThispage

\begin{abstract}
Recent successes in applying reinforcement learning (RL) for robotics has shown it is a viable approach for constructing robotic controllers. However, RL controllers can produce many collisions in environments where new obstacles appear during execution. This poses a problem in safety-critical settings. We present a hybrid approach, called iKinQP-RL, that uses an Inverse Kinematics Quadratic Programming (iKinQP) controller to correct actions proposed by an RL policy at runtime. This ensures safe execution in the presence of new obstacles not present during training. Preliminary experiments illustrate our iKinQP-RL framework completely eliminates collisions with new obstacles while maintaining a high task success rate.

% end \iffalse
\end{abstract}


\IEEEpeerreviewmaketitle

\section{Introduction}

% lahiatt: in the abstract you seem to focus on unexpected obstacles / elements in the workspace. But here you are introducing the idea of working with a human, and varying layouts being the issue. You should pick one (is it people? layouts? objects?) and be consistent throughout. - In human-robot shared environments robots must be able to adapt their motions in real-time to varying layouts while still maintaining safety and ability to complete complex tasks.  

%Reinforcement Learning (RL) is a powerful tool to solve sequential decision-making tasks in complex domains without the need for knowledge of the workings of the complex system models. In recent years 

Reinforcement Learning (RL) has shown success in generating motions for complex manipulation tasks like opening doors \cite{gu2017deep}, in-hand object manipulation \cite{andrychowicz2020learning}, and multi-arm harvesting ~\citep{li2023multi}.
% Methods in RL do well at learning how to complete complex 
% , long-term 
% manipulation tasks when given enough advanced time to interact with and learn from an environment.
%However, RL still struggles with guaranteeing safety of its final learned motion when executed in an environment very different than it had seen during training. 
However, RL policies are sensitive to the environment in which they are trained \cite{safeRLsurvey22}.  In particular, new obstacles that appear during execution may result in collisions by the RL controller representing safety violations.

Safe RL \cite{safeRLsurvey15} explores how to provide safety guarantees for learned RL policies by constraining a policy during the exploration or execution phases. In the exploration phase of RL, the policy continuously updates based on the rewards observed after taking various actions in different states over a large number of samples. Safe exploration includes an additional condition which penalizes taking an action that leads to a safety-critical situation \cite{simao2021alwayssafe,hunt2021verifiably,yang2021safe,bharadhwaj2020conservative}. However, not all safety critical situations can be known in advance. The resulting policy after safe exploration can therefore still have an unsafe condition if a new safety critical situation appears at runtime. In this work, we focus on execution-time safety which investigates how to continuously ensure safety for a policy as it executes in a workspace where new safety critical situations may occur. An example of this can be new obstacles appear in the workspace at runtime. 
% Many prevalent methods in execution-time safe RL rely on knowing a safe set of actions and/or states to which the policy can recover to if the robot gets close to an unsafe state \cite{recoveryrl, backup_policy1, recoveryrl2,recoveryrl3, kiemel2022learning}.

There are two common approaches for reducing collisions of an RL policy during execution.
%A prevalent method for runtime safety enforcement of RL policies 
The first is shielding \cite{Shielding_Seminal_Paper}, where a shield monitors the policy at runtime and substitutes any proposed action that could lead to a safety violation prior to execution within the environment. Substituting an action involves the shield modifying the proposed action to follow a set of pre-defined safety rules, coming up with a safe alternative action, or making the agent come to a complete stop until it is safe again. Shielding grounds its safety specifications in temporal logic which is powerful to pre-define comprehensive rules to follow throughout the full task execution. However, temporal logic does not easily adapt to consider new safety rules which may appear during runtime. 
% Shields monitor actions using temporal logic constraints to pre-define the desired runtime constraints. Our approach has similar goals in monitoring the policy at runtime for safety adherence but instead aims to build towards being able to input observations of the environment to a controller which could in real-time constrain to new obstacles.
% monitors actions output from a policy and replaces any proposed actions which could lead to a safety violation.
% Learning phase safety blocks off dangerous regions of the space a priori resulting in a policy that conservatively avoids a set of known dangerous regions. However, the full variability in an environment is hard to fully predict in advance. Deployment time safety can be thought of using ...

The second approach explores how adding control methods to a policy can ensure adherence of new constraints at runtime. An example of a new constraint is a new, unexpected obstacle seen at runtime which must be checked and corrected for in real-time. These control methods take in a set of points from the policy and make corrections in real-time to form a trajectory that can avoid constraint violations.
% constraint (and therefore safety) violations. 
% However, such control methods necessitate accurate models of the environment to guarantee this and are best for short-term planning of tasks. However, while such approaches are successful at correcting trajectories in the short term they do not consider the higher-level, long term goal of the problem. As an example the controller may be able to check and correct for the next few steps in a given trajectory but not know the ultimate goal for the agent is to reach a certain object in the workspace.
A popular example of this is learning-based Model Predictive Control (MPC). MPC is a trajectory optimization framework using simple cost functions to define high-level task goals while accounting for constraints on the system and system dynamics \cite{tassa2012synthesis}. MPC continuously solves its optimization problem over a short horizon then takes the first proposed action and repeats the process based on observations on how the system changed. MPCs are limited by the complexity of tasks for which the optimal control problem can be solved on without heavily relying on approximations in the cost function \cite{hewing2020learning}. Learning-based MPC explores the best way to combine MPC (which excels at safety through constraint satisfaction) with learning-based controllers (which excel at optimizing overall performance for complex tasks) \cite{hewing2020learning}. 

% MPC for safe learning \cite{hewing2020learning} uses a learning system to optimize the overall cost while at execution the controller handles the safety through constraint satisfaction as needed. In this method, at every time step MPC verifies safety by computing a safe backup trajectory that modifies the learning output as little as possible.... \cite{wabersich2021predictive}

Our approach in this paper is a hybrid approach similar to learning-based MPC. We leverage a prior algorithm called iKinQP (inverse Kinematics Quadratic Programming) \cite{ashkanazy2023collisionfree}, which is a simplified, lightweight version of MPC that treats the system as a first order linear dynamic system. To combine iKinQP with RL, we
%having the robot's underlying control laws instead handle any non-linearities. Such control methods do very well at constructing trajectories and ensuring their adherence to hard constraints. %However unlike RL, such controllers cannot parse sparse, simple task objectives to complete complex tasks.
% \footnote{Full discussion on related works in Appendix~\ref{related_works}} 
%We propose a method which provides online safety corrections to a pre-trained policy using a lightweight, simplified variant of a Model Predictive Controller called inverse Kinematics Quadratic Programming (iKinQP) controller \cite{ashkanazy2023collisionfree}. This combined system we call iKinQP-RL.
% This method combines an inverse Kinematics Quadratic Programming (iKinQP) controller \cite{ashkanazy2023collisionfree} with RL which we call iKinQP-RL.
% However, it is not always intuitive to combine these two approaches effectively in tasks within home environments. Therefore, we propose a method which adapts pre-trained Reinforcement Learning policies to new environments using a trajectory tracking controller to prune away safety and other constraint violating actions. 
% \textcolor{blue}{Home, but can also apply in factories too!}
% This paper proposes an inverse kinematics quadratic programming reinforcement learning (iKinQP-RL) framework.
first train an RL policy to reach a pre-specified goal region.
% plans long-term trajectories to a pre-specified goal region.
Then at runtime, iKinQP \cite{ashkanazy2023collisionfree} corrects actions suggested by the RL policy to ensure safety while still accomplishing the goal.
% This process is illustrated in Fig.~\ref{fig:ikinqp_rl_wkflow}. 
Safety here refers to guaranteeing static collision obstacle avoidance, robot self-collision avoidance, and adherence to the robot's joint position and velocity limits. Our hybrid approach here leverages the complex task learning from RL and safety corrections from iKinQP.

% we are leveraging the XX of RL and the YY of iKinQP

\textbf{Contributions:} We develop a hybrid approach called iKinQP-RL that, in real-time, corrects trajectories from pre-trained RL policies to eliminate collisions with any obstacles and ensures adherence to joint limits.  
%when deployed to new environments using an iKinQP controller to prune away actions violating safety at runtime. Our preliminary evaluations demonstrate iKinQP-RL's ability to ensure safety while maintaining a high task success rate. 
We demonstrate the effectiveness of iKinQP-RL on a reach task. The reach task has the goal of having a manipulator (holding a peg object) get to a location over a target region. At runtime, we test the ability of iKinQP-RL to correct the pre-learned reach policy with a new block obstacle appearing in different locations at runtime.
Preliminary results demonstrate that iKinQP-RL eliminates collisions while maintaining a high task success rate. Whereas using RL alone has high collision rates when completing the same tasks.
% lahiatt: Good rewrite. You may also want to include that RL alone has a high collision rate.

%\textit{mention of safe RL tie-in?}
% different collision objects, namely a static block and static human. 
% We ground our work against both a baseline of reinforcement learning and  \citeauthor{thumm2023human}'s safety shield in human-robot gym \cite{thumm2023human}. 
% We further exemplify our system's generalizability through showing how our approach can be easily used with another robotic arm with similar success rates. 
% \textcolor{blue}{\textbf{Rework this statement at the end!}}

\section{Background}

% Our system is based on a combination of RL and trajectory tracking controls. The fundamentals of each of these methods is described in detail in the following sections. Then in Section~\ref{technical} we discuss how all of these techniques come together to form our combined iKinQP-RL system for safety enforcement.

% \subsection{Reinforcement Learning} \label{rl}

Our hybrid approach leverages two existing algorithms: a reinforcement learning algorithm and the iKinQP \cite{ashkanazy2023collisionfree} algorithm.  

\textbf{Reinforcement Learning (RL)} 
We can define the RL problem as an infinite-horizon Markov Decision Process (MDP), $M = (\mathcal{S},\mathcal{A},p,r)$ with horizon H. This MDP is made up of a continuous, fully observable state space $\mathcal{S}$, a continuous action space $\mathcal{A}$, an unknown state transition probability $p: \mathcal{S} \times \mathcal{A} \times \mathcal{S} \rightarrow [0,\infty)$, and a bounded reward $r: \mathcal{S} \times \mathcal{A} \rightarrow [r_{min}, r_{max}]$ for each transition \cite{haarnoja2018soft}. The unknown transition probability $p$ is the probability density of the next state $s_{t+1} \in \mathcal{S}$ given the current state $s_t \in \mathcal{S}$ and action $a_t \in \mathcal{A}$. The objective of the Reinforcement Learning agent is to learn a policy $\pi(a_t|s_t)$ that maximizes its expected reward by choosing the best action to take given the current state of the world.

\textbf{Inverse Kinematics Quadratic Programming (iKinQP)} \cite{ashkanazy2023collisionfree} finds a trajectory adhering to constraints when given a start and desired state. These states are defined by a joint position, joint velocity, and a constraint on the amount of time given to get to that state from the previous state. To find this trajectory, iKinQP optimizes a cost function that considers a set of soft and hard constraints. The soft constraints include maintaining low tracking error, low integrator drift, and smooth trajectories. These soft constraints help ensure a precise and smooth trajectory deviating as little as possible from the desired next location. 
% The soft constraints include maintaining low tracking error, low integrator drift to prevent overshoot, and smooth trajectories to minimize jerk. These soft constraints help ensure a precise and smooth trajectory deviating as little as possible from the desired next location. 
Further, the trajectory must adhere to hard constraints of obstacle avoidance as well as joint position/velocity limit adherence.

The collision avoidance constraint requires the robot stay a minimal collision buffer distance $d\textsubscript{coll\_buff}$ away from other robots in the space, including avoiding self-collisions $d\textsubscript{robot,robot}$. It also has the same requirement for avoiding environmental objects  $d\textsubscript{robot,env}$. The collision checking is done for simplified 3D geometry models of the robot and objects, rather than their full mesh models.

\begin{figure}[t]
    \centering
    \includegraphics[width=0.8\columnwidth]{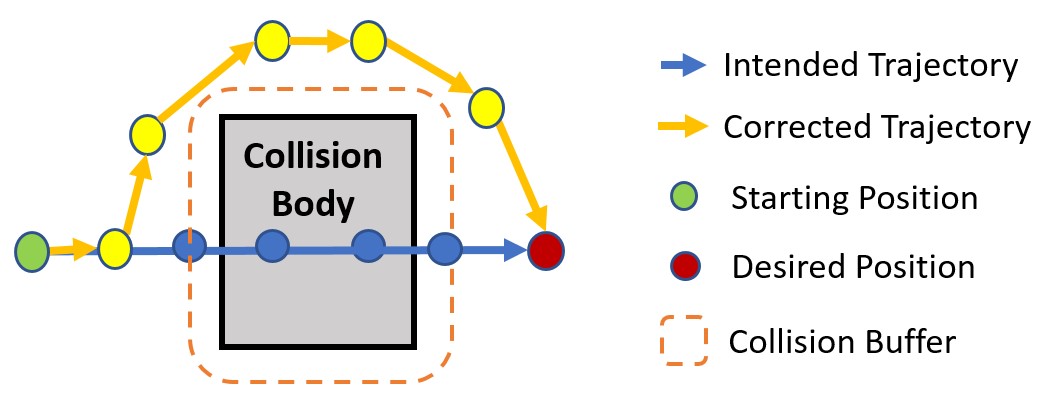}
    \caption{iKinQP collision corrections for simplified 2D example. }
    % \frommak{figures should generally appear at the top or bottom of a column or they create strange gaps in the prose; i will fix this throughout.}}
    \label{fig:iKinQP_collision_avoidance}
\end{figure}

A simplified 2D example of the process showcasing the arm moving safely to reach a goal despite an object in the way is shown in Figure~\ref{fig:iKinQP_collision_avoidance}. iKinQP is given a start (green) and desired final position (red) as input. First iKinQP will greedily interpolate a trajectory (blue) between the two positions ignoring any collision obstacles on the way (gray). Then, iKinQP will check and correct any positions on the way using the collision avoidance constraint resulting in a corrected trajectory (yellow).
% Suppose while trying to go from a starting position (green) to the final position (red) input to iKinQP, the proposed greedy trajectory (blue) will collide with an obstacle on the way (gray). In this case, iKinQP will use the collision avoidance constraint to propose a new trajectory (yellow) that avoids collisions with the obstacle but still ends up as close to the target point as it can.
% meets the overall objective.  
Full details of the iKinQP algorithm are in Appendix~\ref{Appendix_iKinQP}. 
% \frommak{probably need to modify the figure to show the final objective that resides outside and to the right of the gray box.  Otherwise the problem is infeasible.}

%as it is lightweight, operates in real-time, and generates smooth, collision-free trajectories. It can handle either joint space or end effector (EEF) space inputs.  In our problem, we use the joint space to have the most control over the robot movements (More details are in Appendix~\ref{joint_position}). 
% The case of using EEF inputs uses the inverse kinematics (iKin) component of the algorithm, as it finds joint positions given the position of the end effector.
% Therefore, 
\iffalse
From mak: I think this paragraph is not needed given the detail below

iKinQP receives as input the current and the next desired joint position, joint velocity, and timing constraints. Timing constraints define the amount of time to go from the initial to desired joint position. Further, timing is used to determine how 
the desired motion can be broken down into a series of intermediate points. Each point is then checked and corrected for a series of pre-defined soft and hard constraints at every step. This is done using methods in Quadratic Programming (QP) which can solve constrained optimization problems for a multivariate quadratic function. 
\fi %end of \iffalse

% This collision avoidance constraint can be formalized as: $ d_{robot,b} =  \lVert{q_1 - q_0}\lVert, ~ b \in \{robot,env\}$. The iKinQP algorithm does collision detection using NTCD \cite{Grieverheart2017} which is based off of shortest distance calculations between pairs of elementary collision volumes. The collision checking is done for simplified 3D geometry models of the robot and objects, rather than using full mesh models.

% \vfill

% The first step is to check if the current joint position is far enough away from colliding with the robot or any environment obstacles (line 6). If this is the case then the current joint position is set as the failsafe joint position $q_{last\_safe}$ (line 7). More formally, we save a new $q_{last\_safe}$ when $min([d_{robot,robot}(q_0), d_{robot,env}(q_0))] > collision\_thres + \textnormal{\emph{small\_buffer}} > 0$ ($\alpha= collision\_thres + small\_buffer$). However, when the arm is close to collision we instead choose a pre-defined conservative failsafe position $q_{cfs}$ high above the workspace which is guaranteed to be safe (line 8-10). The $q_{fs}$ is chosen to correspond with the section of the table (left, center, or right) to which the gripper is closest 

\section{The iKinQP-RL Approach} \label{technical}

\begin{figure*}[t]
    \centering
    \includegraphics[width=0.7\textwidth]{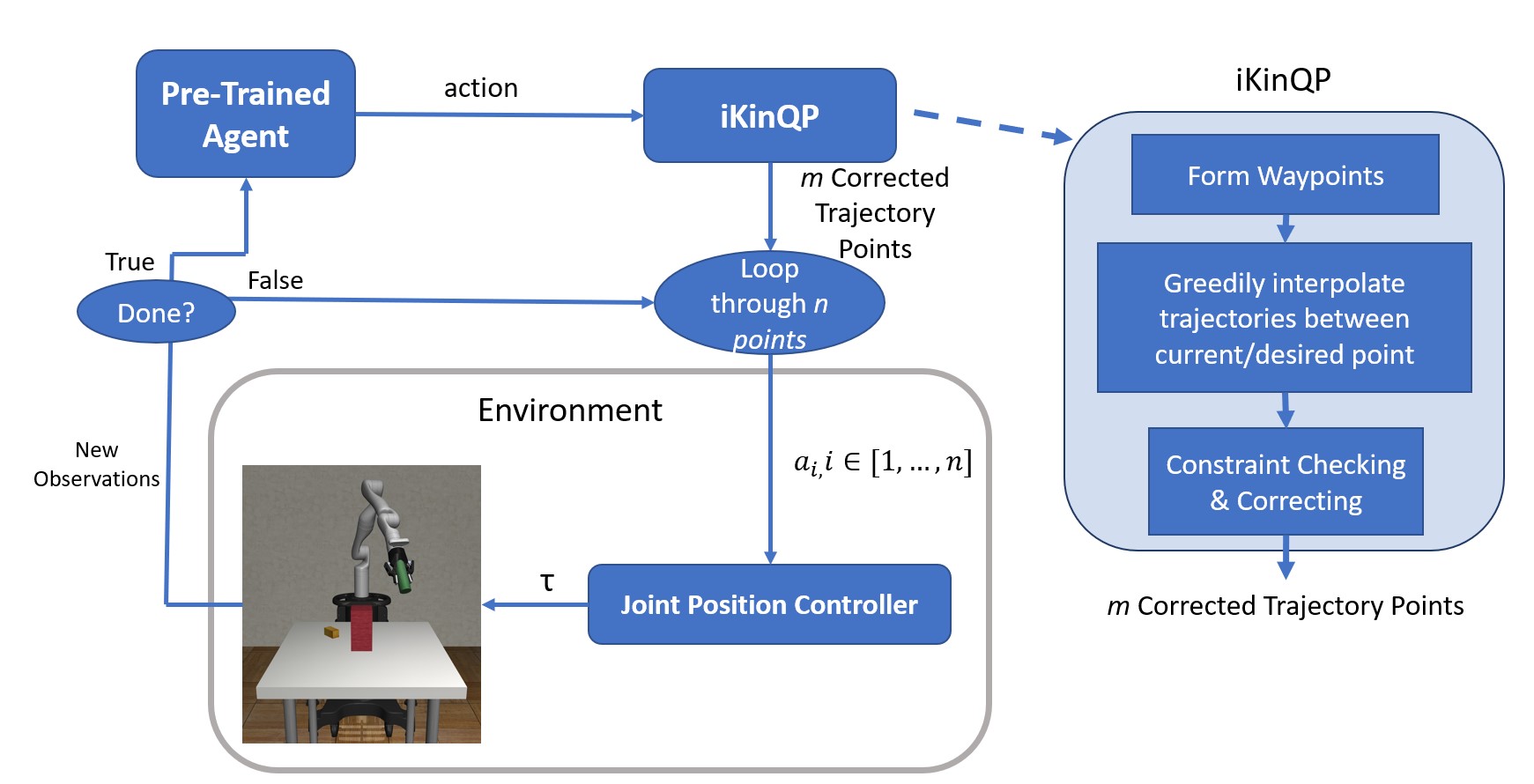}
    \caption{Proposed iKinQP-RL Framework}
    % {\textbf{iKinQP-RL Framework:} At every step, a pre-trained agent outputs the next action to take as well as environment observations which are then fed through the inverse Kinematics Quadratic Programming (iKinQP) controller. iKinQP uses the information from the agent to form a set of Waypoints then greedily interpolates trajectories between the initial and desired position. Each intermediate point of the trajectory is checked and corrected for constraint violations resulting in a set of $m$ safe trajectory points. A subset ($n<m$) of these points are looped through the environment where a joint position controller changes the points to torques ($\tau$) used to move the robot in the environment. This is continued until `done' is triggered which corresponds to a set of termination conditions whereby the process begins again.}
    \label{fig:ikinqp_rl_wkflow}
\end{figure*}

Fig.~\ref{fig:ikinqp_rl_wkflow} illustrates the iKinQP-RL framework design. First, a pre-trained RL agent outputs an action which can be represented in joint space\footnote{iKinQP provides support for end effector (EEF) actions with iKinQP's optional inverse kinematics module. For the sake of our study we focus solely on the joint space representations as we wanted more control over each joint. This helps in checking for safety of each joint, whereas EEF controllers are faster but prone to multiple joint position solutions making them harder to check for safety.} of the next desired goal position/orientation of the agent. This action represents a delta between the desired final joint position ($q_1$) and initial joint position ($q_0$), $a_{q} = q_1 - q_0$. Then, a modified iKinQP\footnote{The modified iKinQP algorithm is explained in more detail in Appendix~\ref{Appendix_iKinQP}.} checks and corrects the action (if necessary) forming a set of $m$ corrected trajectory points, see Fig.~\ref{fig:iKinQP_collision_avoidance} for an example of this process. The modified iKinQP accounts for the possibility of a starting position input to iKinQP falling within the collision buffer of an obstacle (0.015m). If this ever happens iKinQP falls back on a failsafe position that is guaranteed to be safe\footnote{More details on the positions chosen for falling back on when the starting position in iKinQP is within the collision threshold can be found in Appendix~\ref{failsafe_explained}.}. We send $n$ $<$ $m$ of the corrected trajectories points from iKinQP to the environment\footnote{We found that we can only send a subset of the \textit{m} corrected points directly to the environment before querying the policy again for a new action (line 3). The choice of $n$ can effect efficiency and safety of the corrected policy. Full details on finding the right $n$ can be found in Appendix~\ref{grounding_n}}. Each action sent to the environment uses the joint position controller to convert each action into the next desired robot torque ($\tau$) sent to the robot.

\begin{algorithm}[t]
% \caption{iKinQP-RL($\pi(a_t|s_t)$, $t_1 = 0.5$, $\dot{q}_1 = 0$)}\label{alg:RL-iKinQP}
\caption{iKinQP-RL($\pi(a_t|s_t)$)}\label{alg:RL-iKinQP}
\begin{algorithmic}[1]
% \For{all tested episodes}
\While{episode step number $<$ max time steps}
    \State Observe current position $q_0$ and velocity $\dot q_0$ 
    % \If{far from collision obstacles}
    %     \State $q_{last\_safe} \gets q_0$
    %     \Else{}
    %     \State $q_{last\_safe} \gets$ conservative safe position $q_{cfs}$
    % \EndIf{}
    \State $a_{t+1} \gets \pi(a_t|s_t)$  \Comment{Get next RL action} 
    % Get from model next action to take in the \textbf{env}
    % \State Clip a_t \& use to calculate next joint position ($q_1$)
    \State $q_1 \gets q_0 + clipped(a_{t+1}) $ 
    % \State $WP_0 \gets [q_0 \ \dot{q}_0 \ t_0], \enspace WP_1 \gets [q_1 \ \dot{q}_1 \ t_1]$
    % \State $WP_1 \gets [q_1 \ \dot{q}_1 \ t_1]$
    % \State Form Waypoints $WPi=[q_i \ \dot{q}_i \ t_i], i \in [1,2]$
    
    % \State $\mathbf{q_c} \gets$ iKinQP($WP_0, WP_1, q_{last\_safe}) $ 
    \State $\mathbf{q_c} \gets$ iKinQP($q_0,  \dot{q}_0, q_1, \dot{q}_1=0, q_{last\_safe}) $ 
    \For{$i = 0,1,...,n$} \Comment{$n=\lceil \frac{m}{2} \rceil, m = size(q_c)$}
        \State Observe current position $q_0$
        \State Corrected action $a_{t+1,c} \gets \mathbf{q_c}[i] - q_0 $ 
        % \State Use \textbf{$q_c$}[i] to calculate new action to take
        \State $obs, done \gets \ensuremath{\textnormal{Step through \textbf{env} with }} a_{t+1,c}$ 
        % \State $q_1 \gets obs[`joint\_pos']$ \Comment{Observe position after step through \textbf{env}}
        \If{$done$ True or $obs$ position close to colliding} 
            \State Stop stepping through $\mathbf{q_c}$ 
        \EndIf
    \EndFor
\EndWhile
% \EndFor
\end{algorithmic}
\end{algorithm}

% \subsection{iKinQP-RL Algorithm}
% $\textnormal{WP}_0$ is a waypoint.
Algorithm~\ref{alg:RL-iKinQP} shows the full iKinQP-RL process in detail. The inputs to the system is the trained policy ($\pi(a_t|s_t)$).
Until we reach the maximum number of time steps for an episode (Line 1), corresponding to either completing a task or running out of time to complete the task, we go through the following steps. We first observe the current joint position ($q_0$) and velocity ($\dot q_0$) (Line 2) to use as inputs to iKinQP.
% to help determine which failsafe position ($q_{last\_safe}$) to use (lines 4-7). The choice of whether the $q_{last\_safe}$ is the current joint position (lines 4-5) or a conservative safe position (lines 6-7).
% A failsafe joint position can be defined as a position that will always be safe enough to fall back on using as the desired position to go towards (instead of what is output from the policy) if the arm gets too close to colliding with an obstacle. The first step is to check if the current joint position is far enough away from colliding with the robot or any environment obstacles (line 6). If this is the case then the current joint position is set as the failsafe joint position $q_{last\_safe}$ (line 7). More formally, we save a new $q_{last\_safe}$ when $min([d_{robot,robot}(q_0), d_{robot,env}(q_0))] > collision\_thres + \textnormal{\emph{small\_buffer}} > 0$ ($\alpha= collision\_thres + small\_buffer$). 
% However, when the arm is close to collision we instead choose a pre-defined conservative failsafe position $q_{cfs}$ high above the workspace which is guaranteed to be safe (line 8-10).
% The $q_{fs}$ is chosen to correspond with the section of the table (left, center, or right) to which the gripper is closest \footnote{Refer to Appendix~\ref{failsafe_explained} for more information on deriving these failsafe joint positions.}.
The trained policy model $\pi(a_t|s_t)$ is then used to generate the next action to take in the environment given the current environment state (Line 3). These actions are clipped so that iKinQP can better handle the collision cases. Specifically, actions output from the policy can be as large as $\pm 1 m$ in size, but iKinQP actions are $\pm 0.2 m$. Clipping, therefore, allows us to more frequently check for collision cases from the policy and is tune-able if a less conservative safety check is desired. We can determine the next desired joint position ($q_1$) using the observed current joint position ($q_0$) and the clipped action (Line 4). We assume the velocity at the next desired joint position ($\dot{q}_1$) to be zero to be the most conservative. In doing so, we state that the arm should always be slowing down as if always approaching a possible collision state\footnote{In future works we will explore making this parameter less conservative so that the arm only slows down when the desired point is near an obstacle.}. We feed the current joint position ($q_0$), current joint velocity ($\dot{q}_0$), next desired joint position ($q_1$), next desired joint velocity ($\dot{q}_1=0$), and the failsafe position ($q_{last\_safe}$) to our modified iKinQP (Line 5). This results in a set of \textit{m} corrected intermediate trajectory points ($\mathbf{q}_c$) connecting $q_0$ to $q_1$.

From the $m$ corrected intermediate trajectory points ($\mathbf{q}_c$) we found that using an $n$ of the ceiling of half of the intermediate points $n=\lceil \frac{m}{2} \rceil$ to be best \footnote{Full details can be found in Appendix~\ref{grounding_n}}. 
% We found that this choice helped best balance the trade-off between efficiency and safety based on the small error in position that occurs when stepping through the environment multiple times in a row.
% \footnote{Full details on how we chose $n$ to best balance efficiency and safety can be found in Appendix~\ref{grounding_n}.}. 
For each of point in $\mathbf{q}_c$ from $0$ to $n$ (Lines 6-11), we loop through the following steps. During each iteration of the for loop (Lines 6-11), the state of the robot changes. Therefore, at the start of each iteration of looping through the $n$ points we observe the current joint position (Line 7) to construct the new next corrected action ($a_{t+1,c}$) to send to the environment (Line 8). We then step through the environment with $a_{t+1,c}$ (Line 9). An environment step can be defined as using the joint position controller to convert actions to torques sent to the robot within the simulator environment. 
% Actions here represent deltas between initial and desired positions which also include information added back in on the gripper's state. 
Our connection of iKinQP and RL has all of the safety checks and corrections done before executing a series of steps of corrected points in the environment. However, after each step in the environment there is a possibility of a small error between the desired and achieved position. Therefore, if we find the new observed point is too close to a collision obstacle or if the environment has terminated due to reaching its full reward (done) we terminate the loop through the $n$ corrected points early (Lines 10-11). 
%lahiatt: But the algorithm just says "if done then stop stepping through q_c"...? maybe you're saying line 16 returns done sometimes. But the way this is written is a bit confusing. Honestly, I'd just leave it out. it's a small implementation detail that really isn't necessary here.
% Jun 17, 2024 10:47 AM
% lahiatt: (these last three sentences)

% After looping through all $n$ points if the maximum horizon is reached for the episode then the episode will be terminated (lines 23-25).

% \textbf{I also am getting confused about actions vs. joint positions. So the policy is selecting actions but iKinQP is picking joint positions? Is this a meaningful difference?}

\section{Experiment and Results}

\subsection{Experimental Setup}

\begin{figure}[t]
     \centering
     % \begin{subfigure}{0.25\columnwidth}
     \subfigure[Middle]
     {
         \centering
         \includegraphics[width=0.25\columnwidth]{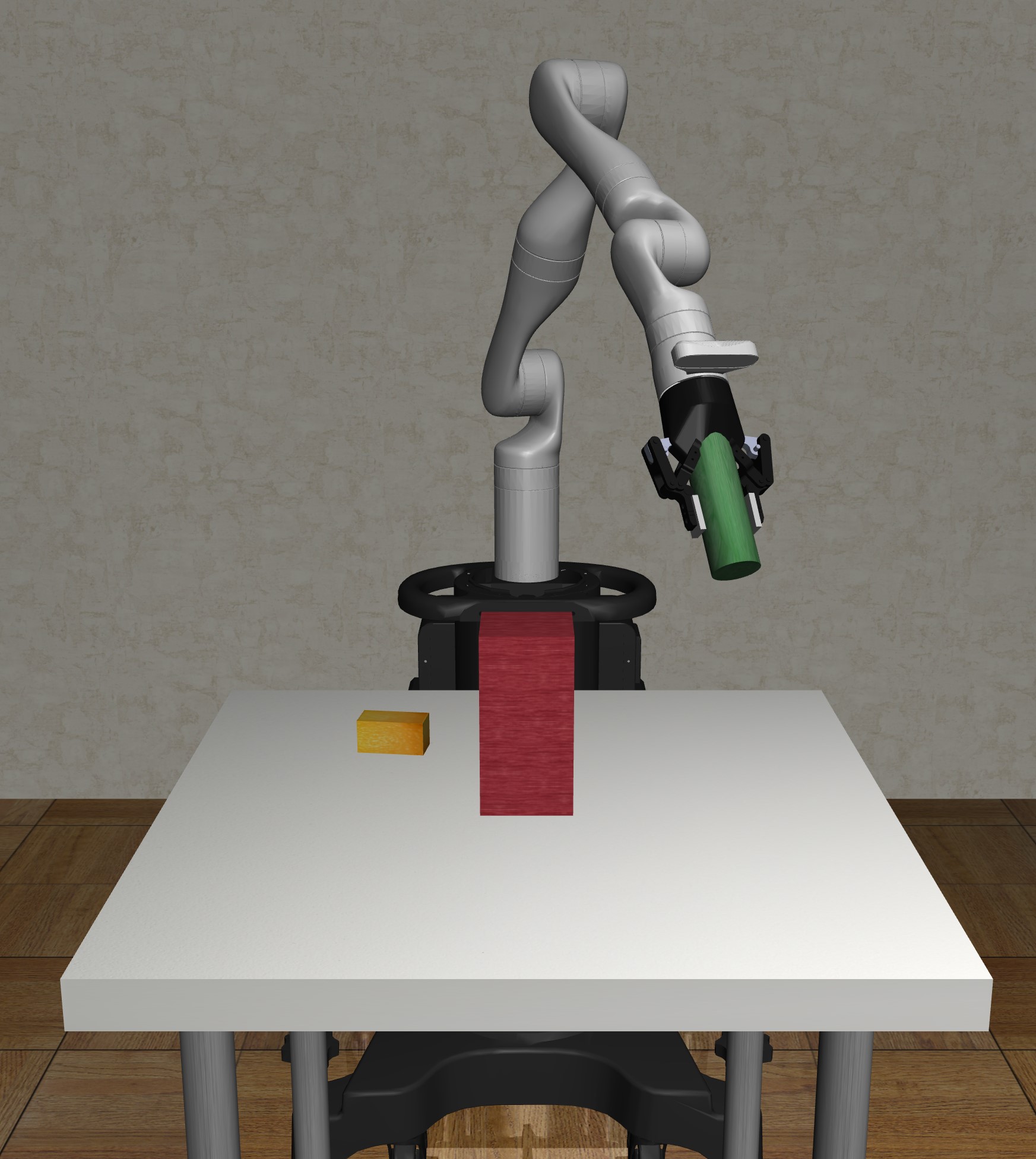}
         % \caption{Block-1-Middle}
         \label{fig:Fixed-Block-1}
     }
     % \end{subfigure}
     \hfill
     \subfigure[Partial-Block]
     % \begin{subfigure}{0.28\columnwidth}
     {
         \centering
         \includegraphics[width=0.28\columnwidth]{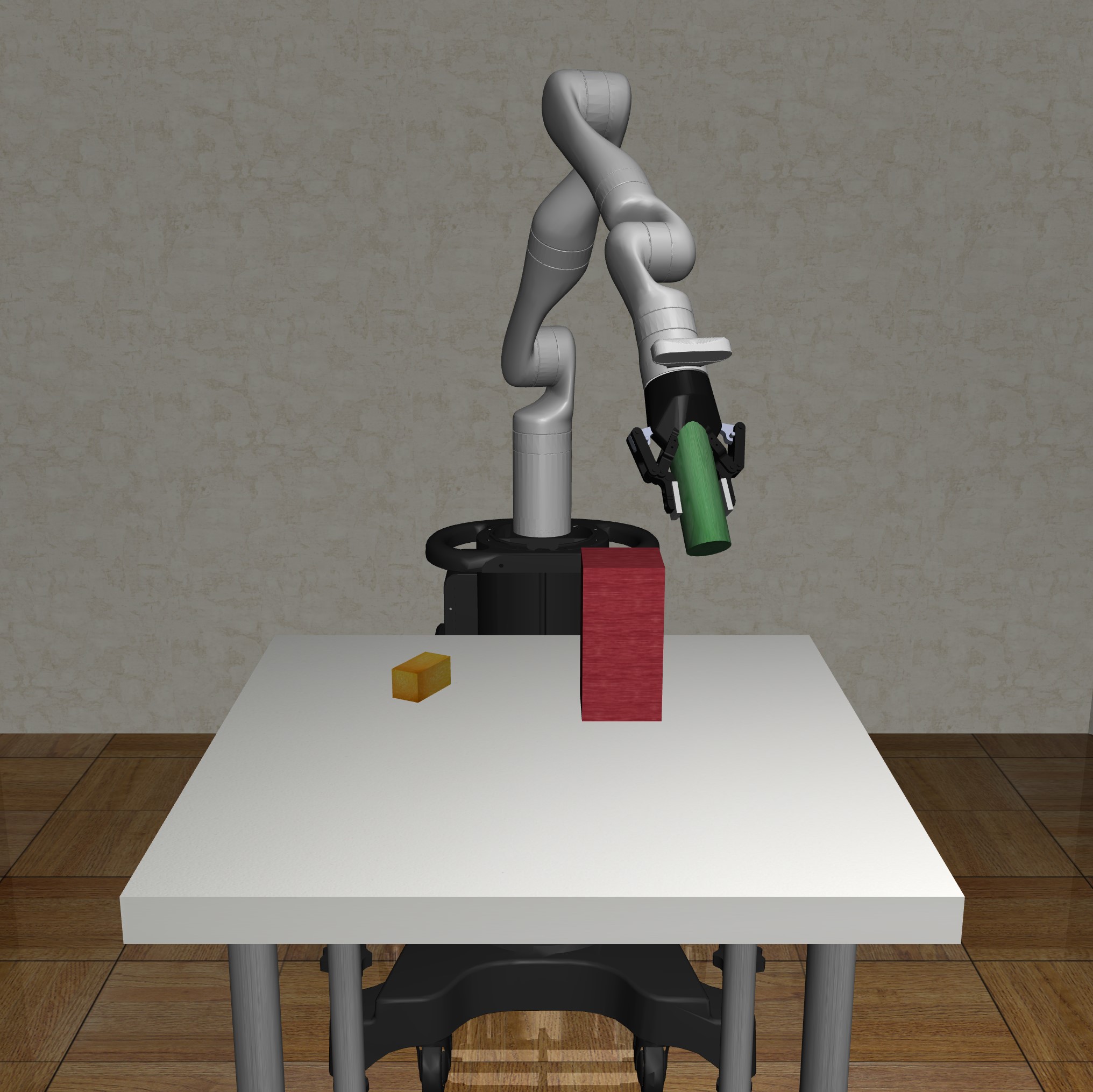}
         % \caption{Block-1-Alt-Loc-1}
         \label{fig:Block-1-Alt-Loc1}
     }
     % \end{subfigure}
     \hfill
     % \begin{subfigure}{0.28\columnwidth}
     \subfigure[Far-Away]
     {
         \centering
         \includegraphics[width=0.28\columnwidth]{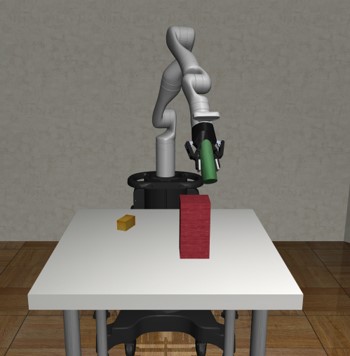}
         % \caption{Block-1-Alt-Loc-2}
         \label{fig:Block-1-Alt-Loc2}
    }
     % \end{subfigure}
        \caption{Experimental Conditions for Block-1}
        \label{exp-conditions-p1}
\end{figure}

We test our system in the robosuite \cite{robosuite2020} simulation framework, which builds off the MuJoCo physics simulation engine \cite{todorov2012mujoco}. We consider a task where the robot is trained to carry an object to the opposite end of the table. Then, at runtime, a new static block appears on the table in one of three positions making up three experimental conditions. Fig.~\ref{exp-conditions-p1} shows the three experimental conditions considered for the robot reaching for towards a goal region (shown visually as over the yellow block). 
% while avoiding an obstacle not seen during training (red block).
The new obstacle added at runtime for our tests is single red block which can be at one of three locations: middle of the workspace blocking the task completely (Middle), near the task space but not fully blocking it (Partial-Block), or not very close to the task space of the robot (Far-Away). Here, whether the new obstacle blocks the task space is determined by whether the obstacle hinders the ability of the robot to complete the trained reach task.
% We assume the red block was not there during the policy's training, but in execution iKinQP has information on the position and shape of the obstacle.

We train the policy with Soft Actor Critic (SAC) \cite{haarnoja2018soft} because it is the state of the art and is designed to work well in continuous, high-dimensional action spaces. This is done while balancing exploration-exploitation during training. While we chose SAC for our preliminary tests, we designed iKinQP-RL to work with any chosen RL algorithm. We train our SAC policy in Stable-Baselines3 \cite{stablebaselines3} for 500 epochs and 8M time steps\footnote{Full details on the details of the RL experimental setup can be found in Appendix~\ref{RL_Exp_Details}.}. We trained our system on a GPU server with a AMD EPYC 7H12 64-Core Processor CPU, 2.0 T of RAM and the run-time experiments were conducted on a laptop computer with a 4 x 1.90 GHz Intel Core i7-8650U CPU, 7.9 GB of RAM running Windows Subsystem for Linux. iKinQP is implemented in the Julia programming language.
% and the Reinforcement Learning component was written in Python with no multi-threading.

% While for this study we chose to use SAC to train the robot's policy, our system is designed to be generalizable to different policy types.
We use the Kinova Gen3 (Kinova3) robotic manipulator, a 7-DOF robot arm equipped with a two fingered Robotiq85Gripper. The observational space of the policy consists of camera images, force-torque sensor readings, pressure signals from sensors on the robot's fingers, the location of the goal object, the current joint position, and the current joint velocity. 
% Joint sensors give information on the current position and velocity of the robot's joints.  
A joint position controller with an 8D action space is used to control the robot arm. The first 7D represent the 7 DOF for the Kinova3 arm which are delta values from the current state of the robot to the next state. The last dimension is the Robotiq 85 gripper's open/close action.

\subsection{Results} \label{results}

We evaluated iKinQP-RL by comparing its collision rate and task success rate against the baseline of a trained RL policy without using iKinQP. 
The collision cases which had to be avoided were with the table, red block object, and self-collisions. It was not considered unsafe to touch the yellow block. We determined task success based on whether by the end of each episode object carried by the robot is positioned over a region close to or touched the top of the target region (yellow block).
\iffalse
\frommak{this is probably too much detail.  }
% lahiatt: Does this matter? Why don't you just say that robosuite detects collisions and you're counting them?
%lahiatt: I think I mentioned earlier but somewhere (abstract, intro and then again in this section) you should lay out your claims (safer, faster, better, etc.). then here at a minimum you should say how you are measuring those (like in the beginning of the experimental section). Bringing them up as they come up weakens the narrative.
% "safety analysis" term vague/undefined...
For our safety analysis, we investigate the collision rate ($\%$) in each of the 100 episodes we roll out. Collisions come in two forms corresponding to table bodies (tb): $c_{table,j}, j=\{1,..,tb\}$ and object bodies (ob): $c_{object,k}, k =\{1,..,ob\}$. These collision types $c_{table,j}, c_{object,k} \in \{0,1\}$ representing no collisions with the table/object (0) or a collision with the table/object (1). Therefore, the total number of collisions for all collision bodies $n = tb + ob$ can be represented by $c = \sum_{i \in \{1,...,n\} } c_i$. To have guaranteed safety, the combined RL-iKinQP system must have $c=0$ corresponding to a 0$\%$ collision rate.
\fi

We examined the collision rate over 100 episodes when relying only on the trained RL policy without iKinQP (Baseline) to ground the difficulty of the different experimental conditions. Table~\ref{safety-results_paper} shows these results. The Middle environment was the most difficult and had a $100\%$ collision rate, since the block always was in the way of reaching the goal. The Partial-Block was slightly easier having an $86\%$ collision rate, since the block was more out of the way of the task space. Lastly, Far-Away was the easiest environment with a $32\%$ collision rate. Further, we measured the per-task success rate which defines the success of the trained policy, here referring to the ability of the robot to reach its target position. All environments in the Baseline condition maintained a $100\%$ per-task success rate.

% All of these environments in the baseline condition maintained a $100\%$ per-task success rate. The baseline condition refers to using a trained RL policy with no iKinQP corrections. 
% lahiatt: "Baseline condition" -- not a defined term yet (?). Also -- how can it have such a high task success rate when it collides? I'm not sure I understand.

% Jun 17, 2024 12:06 PM
% You: It is still able to reach the goal even if the arm knocks over the block on the way. The policy doesn't know the block is there therefore it will record a success even if the block is knocked over on the way.

% Jun 17, 2024 11:19 PM • Edit • Delete

iKinQP-RL stays outside of the collision buffer of all collision objects\footnote{A visualization of the minimum collision proximity per episode for each experimental condition can be seen in Fig.~\ref{Base_Coll_Prox}, Fig.~\ref{Alt1_Coll_Prox}, Fig.~\ref{Alt2_Coll_Prox} in Appendix~\ref{collision_distances}.}. The robot's closest distance to colliding over all experimental conditions tested was 0.0167m, which is outside of the collision buffer (0.015m). Table~\ref{safety-results_paper} shows the collision rate when using iKinQP-RL was $0\%$ across all experimental conditions. This indicates that we are able to ensure safety across all scenarios. Further, the per-task success rate was high for all tasks falling between $83\% - 100\%$. This result indicates that preventing collisions does not significantly hinder system performance.  The only time task success was hindered by preventing collisions was in the Middle case ($83\%$). In this case the block is in the way of reaching the goal position and in some experiments we saw this cause the arm to have trouble figuring out a path around the obstacle.

\begin{table}
\centering
\resizebox{\columnwidth}{!}{
\begin{tabular}{|c||r||r|r|} 
\hline
 & Baseline Collision \% & iKinQP-RL Collision\% & iKinQP-RL Task \%  \\ \hline % & Time (ms) \\ \hline
Middle & $100$ & \textbf{$0$} & $83$ \\ \hline % & $102$ \\ \hline
Partial-Block & $86$ & \textbf{$0$} & $100$ \\ \hline % & $101$ \\ \hline
Far-Away & $32$ & \textbf{$0$} & $100$ \\ \hline % & $102$ \\ \hline
% Block-Rand-1 & $\#$ & $\#$ & $\#s$  \\ \hline
% Block-Mult & $\#$ & $\#$ & $\#$ & $\#s$ & $\#s$  \\ \hline
% Human-1 & $\#$ & $\#$ & $\#$ & $\#s$ & $\#s$  \\ \hline
% Fixed-Fixed-Mult & $\#$ & $\#$ \\ \hline
% 6 & $\#$ & $\#$  \\ \hline
\end{tabular} }

\iffalse
original
\resizebox{\columnwidth}{!}{
\begin{tabular}{|c|c|c|c|c|c|} 
\hline
Experiment & Coll. w/out iKinQP & Coll. Rate & Coll. Prox. & Success Rate ($\%$)  & iKinQP Call Time (s) \\ \hline
Middle & $100\%$ & $0\%$ & $0.0167m$ & $83\%$ & $0.102s$ \\ \hline
Partial-Block & $86\%$ & $0\%$ & $0.0173m$ & $100\%$ & $0.101$ \\ \hline
Far-Away & $32\%$ & $0\%$ & $0.0184m$ & $100\%$ & $0.102s$ \\ \hline
% Block-Rand-1 & $\#$ & $\#$ & $\#s$  \\ \hline
% Block-Mult & $\#$ & $\#$ & $\#$ & $\#s$ & $\#s$  \\ \hline
% Human-1 & $\#$ & $\#$ & $\#$ & $\#s$ & $\#s$  \\ \hline
% Fixed-Fixed-Mult & $\#$ & $\#$ \\ \hline
% 6 & $\#$ & $\#$  \\ \hline
\end{tabular} }
\fi
\caption{Evaluation for each experimental condition: Collision rate without using iKinQP (Baseline), Collision rate using iKinQP-RL (ours), and task success rate for iKinQP-RL (ours) \label{safety-results_paper}}
\end{table}  

% Concluding paragraph of results!!!
Our preliminary experiments show iKinQP-RL ensures safe execution of actions proposed by a pre-trained RL policy. As a comparison, the Baseline in the same experimental conditions resulted in unsafe executions, $32-100\%$ collision rate depending on environmental difficulty. Further, preliminary results show iKinQP-RL balances the performance-safety trade-off well. This meant that when given a pre-trained RL policy with a $100\%$ task success rate, iKinQP-RL prioritized safety without lowering the task success rate significantly. For the experimental conditions where the obstacle blocked the task space less this meant the robot could maintain a $100\%$ task success rate while maintaining safety ($0\%$ collisions). For the hardest task where the obstacle blocked the task space iKinQP-RL always prioritized safety ($0\%$) but still maintained a relatively high per-task success rate ($83\%$). 

\section{Conclusion} 
\label{sec:conclusion}

We presented the inverse Kinematics Quadratic Programming - Reinforcement Learning (iKinQP-RL) framework. iKinQP-RL corrects a Reinforcement Learning (RL) policy in real-time for collision avoidance and joint limit adherence while maintaining smooth, precise trajectories using a modified version of the iKinQP controller. Our preliminary experiments demonstrated iKinQP-RL's ability to balance safety and performance for a manipulator completing a reach task with a new block obstacle at runtime. Future work will explore speeding up our iKinQP-RL approach as well as applying iKinQP-RL to additional tasks like pick-and-place or shared environments with a human user. We also plan to conduct comparisons to shielding approaches to further ground the benefits of our approach.

\section*{Acknowledgments}

This research was supported by the Office of Naval Research. The views and conclusions contained in this document are
 those of the authors and should not be interpreted as necessarily representing the official policies, either expressed or
 implied, of the US Navy. 
 % ONR mention!

%% Use plainnat to work nicely with natbib. 

\bibliographystyle{plainnat}
\bibliography{references}

% \appendix
\begin{appendices}

\section{Experimental Parameters} \label{Appendix_A}

Table~\ref{params} below shows the experimental parameters used in developing iKinQP-RL.

\begin{table} [H]
\centering
{
\begin{tabular}{c c c} 
\hline
Parameter Name & Variable & Value  \\ 
\hline
Joint Position & q &   \\ 
Joint Velocity & $\dot{q}$ & \\
% waypoint & WP & $[q \ \ \dot{q} \ \ t]$ \\ 
% Time to get to WP & t &  \\
% waypoint number & i & i $\in$ [0,1] \\
Action & $a$ & $a = q_1 - q_0$ \\
Distance & d & $\lVert{q_1 - q_0}\lVert $ \\ 
%  &  &  \\ 
% &  &   \\ 
 \hline
\end{tabular}
\resizebox{\columnwidth}{!}{
\begin{tabular}{c c c}  
\hline
Parameter Name & Variable & Value  \\ 
\hline
% Time to first waypoint & $t_0$ &  0.001 \\ 
Time given to get to next desired position (iKinQP) & $t_1$ &  0.5  \\ 
Time step between intermediate points & $\Delta t$ & 0.05 \\
Final Joint Velocity & $\dot{q}_1$ & 0 \\
Collision Buffer & $d_{coll\_buff}$ &  0.015 \\
% Clipped action range & $a_{clip}$ & +/- 0.2   \\ 
Number of points from iKinQP for step  & n &  7 \\
% Safe back-away distance from obstacle & $\alpha = d_{coll\_buff} + buffer $ & 0.15 \\
Proportional Gain of Joint Position Controller & $K_p$ & 50 \\
Derivative Gain of Joint Position Controller & $K_d$ & 0.25 \\
 \hline
\end{tabular}}
% \begin{tabular}{c c}  
% \hline
% Parameter & Value  \\
% \hline
% Replay Buffer Size & 10^6  \\
% Discount Factor \gamma &  0.99  \\ 
% Batch Size & 256 \\
%  &  \\ 
%  &    \\ 
%  &   \\ 
%  &   \\ 
%  \hline
% \end{tabular}
}
\caption{Experimental Parameters \label{params}}
\end{table} 

% \section{Deriving Success Criteria} \label{Appendix_B}

% \begin{equation}
%     \lVert{q_{peg} - q_g}\lVert \leq \epsilon_g
% \end{equation}

% The success of an episode is defined by when the position of the peg ($q_{peg}$) is within some user-defined $\epsilon_g$ distance away from the goal object.

\section{Modified iKinQP Algorithm} 
\label{Appendix_iKinQP}

The modified iKinQP algorithm used in our iKinQP-RL framework is described in Algorithm~\ref{alg:iKinQP}. Internally, the algorithm assumes a given model of the robot arm and environment (including collision obstacles). Further, when going from joint position $q_0$ to $q_1$ within $t_1$ seconds the timing increment assumed here to go between intermediate points is $\Delta t = 0.05$  the collision buffer ($d_{coll\_buff}$) is assumed as $0.015$m.

\begin{algorithm}[H]
\caption{iKinQP($q_0,  \dot{q}_0, q_1, \dot{q}_1=0, q_{last\_safe}$)}\label{alg:iKinQP}
\begin{algorithmic}[1]
% \State Given: Set of \waypoints, $q_{last\_safe}$ \Comment{Note: \waypoint{i} $= [q_i ~ \dot{q_i} ~ t], i \in [0,1]$}
\State Initialize: \textbf{arm} model and environment \textbf{env} using current joint position $q_0$
\State Initialize: $\Delta t, d\textsubscript{coll\_buff}$
% \State Spline Interpolation $\gets True$
\If{$d\textsubscript{robot,robot}(q_0)$$<$$d\textsubscript{coll\_buff}$  or $d\textsubscript{robot,env}(q_0)$$<$$d\textsubscript{coll\_buff} \textnormal{ }$} \Comment{\textit{Addition for iKinQP-RL}}
    \State $q_1 \gets q\textsubscript{last\_safe}$
    \State $a_q \gets q_1 - q_0$
\EndIf
\State $\textnormal{Initialize constraints for Optimizer using} \Delta t, d\textsubscript{coll\_buff},$  $\textbf{arm}, \textbf{env}, q_0$
\State Initialize Optimizer with initial conditions
\If{Interpolate?} 
    \State Create spline between $q_0$ and $q_1$
\EndIf
\For{$t=t_0$ to $t_1$ in $\Delta t$ increments} 
    \If{Interpolate}
        \State Make intermediate spline points between $q_0$ \& $q_1$
        \State $\textnormal{desired\_state} \gets$ spline  point at $t$
        \Else{}
        \State $\textnormal{desired\_state} \gets q_1$
    \EndIf
    \State $q_i, \dot{q_i} \gets$ Solve Optimizer
    \State Save $(q_i, \dot{q_i}, t)$
\EndFor
\State Using $q_i's$, construct full trajectory $\textbf{q}$
\State Using $\dot{q_i}'s,$ construct full trajectory of joint velocities $\dot{\textbf{q}}$
% \RETURN{} \textbf{q}
% \State Return \textbf{$q$}
\end{algorithmic}
\end{algorithm}

\section{Grounding Choice of $n<m$ Intermediate Points Taken from iKinQP Output} \label{grounding_n}

For our chosen set of parameters, $t_1=0.5$ and $dt=0.05$, there are 10 points after the first position leading to the final desired position. Therefore, we can choose up to $m=11$ points to send from iKinQP to the environment before querying the policy for another action. We used the Middle environment to ground our testing, where we first did a quick 20 episode test of how the safety and efficiency of the rollout was when using different values of $n$ (see Table~\ref{tab:Intermediete_Point_Investigation} below).

\begin{table}[h]
    \centering
    \begin{tabular}{|c|r|r|r|}
        \hline
        $\#$ Points Sent (n) & iKinQP-RL $\%$ & Task $\%$ & Total Eps Time (s) \\
        \hline
        4 & 0 & \textbf{0} & --- \\
        \hline
        5 & 0 & 70 & 46.655 \\
        \hline
        6 & 0 & 95 & 29.41 \\
        \hline
        7 & 0 & 90 & 21.71 \\
        \hline
        8 & 0 & 95 & 16.16 \\
        \hline
        9 & $ \textbf{20} $ & --- & --- \\
        \hline
        11 (all) & $\textbf{15}$ & --- & ---  \\
        \hline
    \end{tabular}
    \caption{Number of Points to Send (n) from iKinQP to Environment before querying policy again tested for amount of collision, timeouts, and total time for episode completion for 20 episodes}
    \label{tab:Intermediete_Point_Investigation}
\end{table}

If too few of the \textit{n} points are sent then the success rate can reduce down to $0\%$. Further, if too many of the \textit{n} points are sent then collisions start to occur. Therefore, the key is to find a balance between accuracy (high success rates) and safety (zero collision rate). The most promising candidates for was to use $\lceil \frac{m}{2} \rceil \pm 1$ corresponding to $6-8$ points. We then conducted a second test of extending the trials to 100 episodes with the candidates $n=6,7,8$ testing additionally the average closest distance to colliding with an obstacle (Collision Proximity). Through this, we found that overall $n=7$ (corresponding to $n=\lceil \frac{m}{2} \rceil$)  best balanced the task success with timing and collision proximity Table~\ref{tab:Intermediete_Point_Investigation_100}.

\begin{table}[h]
\centering
\resizebox{\columnwidth}{!}{
\begin{tabular}{|c|r|r|r|r|} 
\hline
        $\#$ Pts & Collision $\%$ & Task $\%$ & Total Eps Time (s) & Collision Proximity (m) \\
        % & Action after \textit{n} (m)  \\
        \hline
        6 & 0 & 84 & 30.65 & 0.0183 \\
        % & 0.003116 \\
        \hline
        7 & 0 & 88 & 24.88 & 0.0167 \\
        % & 0.004496 \\
        \hline
        8 & 0 & 93 & 21.88 & 0.0152 \\
        % & 0.00636 \\
        \hline
\end{tabular}}
\caption{Number of Points to Send (n) from iKinQP to Environment before querying policy again tested for amount of collision, task success, average total time for episode completion, and closest collision proximity over all episodes. \label{tab:Intermediete_Point_Investigation_100}}
\end{table} 

\section{Visualizing Closest Distance to Colliding Per Episode} \label{collision_distances}

% \begin{figure}[h]
%     \centering
%     \includegraphics[width=.45\textwidth]{Figures/Closest_Dist_to_Colliding_100_exps_5_22_24.jpg}
%     \caption{Closest distance to colliding in each episode of system test for Fixed-Block-1 Experiment. Orange line at bottom shows where the collision threshold begins.}
%     \label{fig:closest_dist_to_colliding}
% \end{figure}

\begin{figure}[H]
     \centering
     % \begin{subfigure}[b]{0.3\columnwidth}
     \subfigure[Middle]
     {
         \centering
         \includegraphics[width=0.3\columnwidth]{Figures/Fixed-Block-1-Prob1.jpg}
         % \caption{Block-1-Middle}
         \label{fig:Fixed-Block-1}
     % \end{subfigure}
     }
     \hfill
     % \begin{subfigure}[b]{0.67\columnwidth}
     \subfigure[Closest distance to colliding in each episode of system test for Middle Experiment. Orange line at bottom shows where the collision threshold begins.]
     {
         \centering
         \includegraphics[width=0.75\columnwidth]{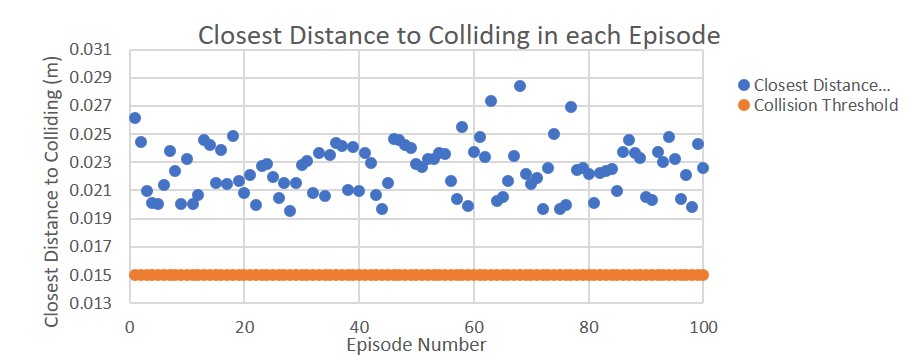}
         % \caption{Closest distance to colliding in each episode of system test for Block-1-Middle Experiment. Orange line at bottom shows where the collision threshold begins.}
         \label{fig:closest_dist_to_colliding}
     % \end{subfigure}
     }
     
        \caption{Red block location chosen to directly block the path of the arm in getting to the goal region over the yellow block.}
        \label{Base_Coll_Prox}
\end{figure}

\begin{figure}[H]
     \centering
     % \begin{subfigure}[b]{0.3\columnwidth}
     \subfigure[Partial-Block]
     {
         \centering
         \includegraphics[width=0.3\columnwidth]{Figures/Block-1-alt-loc1.jpg}
         % \caption{Block-1-Alt-Loc-1}
         \label{fig:Block-1-Alt-Loc1}
    }
     % \end{subfigure}
     \hfill
     % \begin{subfigure}[b]{0.67\columnwidth}
     \subfigure[Closest distance to colliding in each episode of system test for Partial-Block Experiment. Orange line at bottom shows where the collision threshold begins.]
     {
         \centering
         \includegraphics[width=0.75\columnwidth]{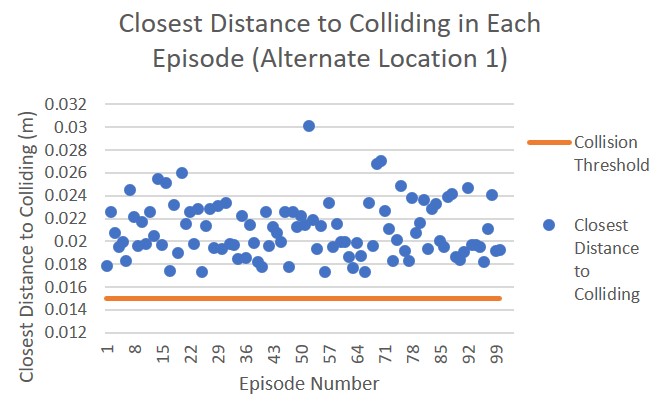}
         % \caption{Closest distance to colliding in each episode of system test for Block-1-Alt-Loc-1 Experiment. Orange line at bottom shows where the collision threshold begins.}
         \label{fig:closest_dist_to_colliding}
    }
     % \end{subfigure}
     
        \caption{Red block location chosen to partially block the path of the arm in getting to the goal region over the yellow block.}
        \label{Alt1_Coll_Prox}
\end{figure}

\begin{figure}[H]
     \centering
     % \begin{subfigure}[b]{0.3\columnwidth}
     \subfigure[Far-Away]
     {
         \centering
         \includegraphics[width=0.3\columnwidth]{Figures/Block-1-alt-loc2.jpg}
         % \caption{Block-1-Alt-Loc-2}
         \label{fig:Block-1-Alt-Loc2}
    }
     % \end{subfigure}
     \hfill
     % \begin{subfigure}[b]{0.67\columnwidth}
     \subfigure[Closest distance to colliding in each episode of system test for Far-Away. Orange line at bottom shows where the collision threshold begins.]
     {
         \centering
         \includegraphics[width=0.75\columnwidth]{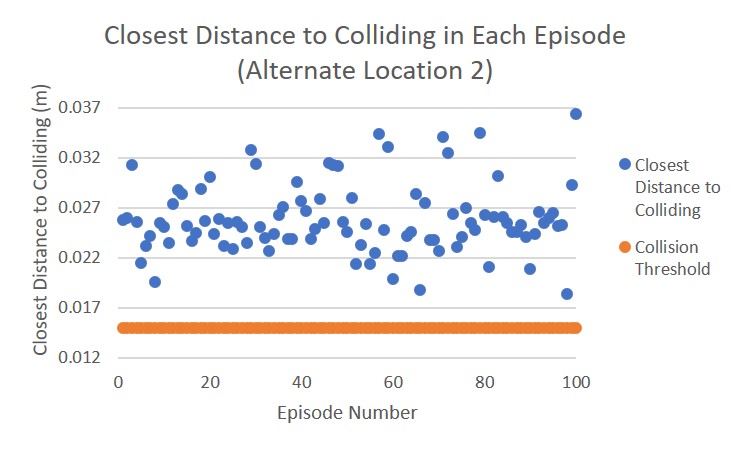}
         % \caption{Closest distance to colliding in each episode of system test for Block-1-Alt-Loc-2. Orange line at bottom shows where the collision threshold begins.}
         \label{fig:closest_dist_to_colliding_alt2}
         }
     % \end{subfigure}
     
        \caption{Red block location chosen to barely block the path of the arm in getting to the goal region over the yellow block.}
        \label{Alt2_Coll_Prox}
\end{figure}

\section{RL Experiment Details} \label{RL_Exp_Details}

The reward function we used is composed of $R=R_{reach}+0.5*R_{y\_align}+R_{close\_gripper}$. $R_{reach}$ (Equation~\ref{reach_reward}) encourages the robot to move the peg object it is holding closer to the target object. $R_{y\_align}$ (Equation~\ref{aligny_reward}) encourages the robot to reach above the cube rather than approaching from the side. This is given a smaller weight than the main goal of reaching the object. $R_{close\_gripper}$ gives a high penalty for the gripper for being open (-100), meaning when the gripper contact points (left and right finger pads) are not in contact with the peg. This is due to the task being assumed to have the gripper always holding the peg. When the reward is 97.5\% or more achieved, the agent will get the full reward and the episode will terminate to prevent over-exploring. 
% We normalize these combined per-step rewards to 1.0 to ensure a maximum per-episode reward of 500.

\begin{equation} \label{reach_reward}
    R_{reach} = 1-tanh(10*d(p,o_k)), d(p,o_k) = \Arrowvert p - o_k \Arrowvert
\end{equation}

\begin{equation} \label{aligny_reward}
    R_{y\_align} = 1-tanh(10*d(y_g,y_{o_k})), d(y_g,y_{o_k}) = \Arrowvert y_g - y_{o_k} \Arrowvert
\end{equation}

Here, the peg position is represented by $p=(x_p,y_p,z_p)$, object $k$ position $o_k = (x_{o_k},y_{o_k},z_{o_k})$, and gripper position $g=(x_g,y_g,z_g)$.

% \section{Experimental Conditions } \label{exp_conditions}

% \begin{figure}[H]
%      \centering
%      \begin{subfigure}[b]{0.25\columnwidth}
%          \centering
%          \includegraphics[width=\columnwidth]{Figures/Fixed-Block-1-Prob1.jpg}
%          \caption{Block-1-Middle}
%          \label{fig:Fixed-Block-1}
%      \end{subfigure}
%      \hfill
%      \begin{subfigure}[b]{0.28\columnwidth}
%          \centering
%          \includegraphics[width=\columnwidth]{Figures/Block-1-alt-loc1.jpg}
%          \caption{Block-1-Alt-Loc-1}
%          \label{fig:Block-1-Alt-Loc1}
%      \end{subfigure}
%      \hfill
%      \begin{subfigure}[b]{0.28\columnwidth}
%          \centering
%          \includegraphics[width=\columnwidth]{Figures/Block-1-alt-loc2.jpg}
%          \caption{Block-1-Alt-Loc-2}
%          \label{fig:Block-1-Alt-Loc2}
%      \end{subfigure}
%         \caption{Experimental Conditions for Block-1}
%         \label{exp-conditions-p1}
% \end{figure}

% \begin{figure}[H]
%     \centering
%      \begin{subfigure}[b]{0.45\columnwidth}
%          \centering
%          \includegraphics[width=\columnwidth]{Figures/Fixed-Block-Mult-Prob1.jpg}
%          \caption{Block-Mult}
%          \label{fig:Fixed-Block-Mult}
%      \end{subfigure}
%      \hfill
%      \begin{subfigure}[b]{0.45\columnwidth}
%          \centering
%          \includegraphics[width=\columnwidth]{Figures/Human-Fixed-1-Prob1.jpg}
%          \caption{Human-1}
%          \label{fig:Human-Fixed-1}
%      \end{subfigure}
%     \caption{Experimental Conditions for Block-Mult and Human-1}
%     \label{fig:exp-conditions-p2}
% \end{figure}

\section{Deriving Failsafe Joint Positions} \label{failsafe_explained}

In the iKinQP-RL Algorithm, the failsafe joint position $q_{last\_safe}$ is defined as the position which the arm will plan to if it is too close to being unsafe. This is defined by the arm being within the collision threshold plus an extra buffer around the object after a step through the environment. Therefore, rather than iKinQP planning a trajectory based on the policy's next action a failsafe position will be used.

\begin{figure}[H]
     \centering
     \subfigure[Failsafe joint position rotated to left side of table]
     {
     % \begin{subfigure}[b]{0.3\columnwidth}
         \centering
         \includegraphics[width=0.285\columnwidth]{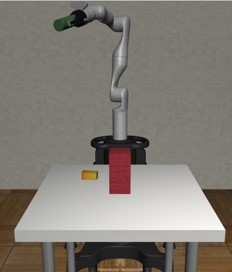}
         % \caption{Failsafe joint position rotated to left side of table}
         \label{fig:Failsafe-Neg}
    }
     % \end{subfigure}
     \hfill
     \subfigure[Failsafe joint position aligned with center of table]
     {
     % \begin{subfigure}[b]{0.3\columnwidth}
         \centering
         \includegraphics[width=0.29\columnwidth]{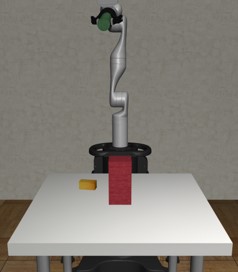}
         % \caption{Failsafe joint position aligned with center of table}
         \label{fig:Failsafe-Zero}
     % \end{subfigure}
     }
     \hfill
     % \begin{subfigure}[b]{0.3\columnwidth}
     \subfigure[Failsafe joint position rotated to right side of table]
     {
         \centering
         \includegraphics[width=0.295\columnwidth]{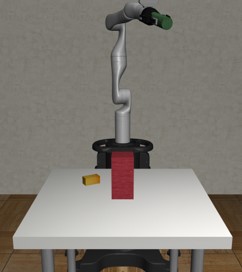}
         % \caption{Failsafe joint position rotated to right side of table}
         \label{fig:Failsafe-Pos}
    }
     % \end{subfigure}
        \caption{Failsafe Joint Positions when close to collision}
        \label{failsafe-positions}
\end{figure}

Whenever the arm is far from colliding from the an object in the environment, the failsafe joint position will be saved as the current joint position. However, if the arm is close to colliding we will instead use a pre-determined failsafe position that puts the arm far above the table rotated towards the section of the table closest to the current position of the arm. This is based on the table being divided into thirds using the width as a reference ($\frac{table\_width}{3}$) resulting in three pre-determined failsafe positions for the arm. The failsafe is then chosen to correspond with the section of the table (left, center, or right) to which the gripper is closest. These failsafe positions can be seen in Fig.~\ref{failsafe-positions}.

\end{appendices}

\end{document}